\documentclass[10pt,twocolumn,letterpaper]{article}
\usepackage[pagenumbers]{cvpr}      

\usepackage[english]{babel}

\definecolor{cvprblue}{rgb}{0.21,0.49,0.74}
\usepackage[pagebackref,breaklinks,colorlinks,allcolors=cvprblue]{hyperref}
\usepackage{multirow}
\usepackage{graphicx}

\def\paperID{6396} 
\def\confName{CVPR}
\def\confYear{2025}

\title{Dust to Tower: Coarse-to-Fine Photo-Realistic Scene Reconstruction from Sparse Uncalibrated Images}

\author{
Xudong Cai$^{1}$ \quad
Yongcai Wang$^{1}$\thanks{Corresponding authors.} \quad
 Zhaoxin Fan$^{2}$\footnotemark[1] \quad 
Deng Haoran$^{1}$ \quad
Shuo Wang$^{1}$ \quad \\
Wanting Li$^{1}$ \quad
Deying Li$^{1}$ \quad
Lun Luo$^{3}$ \quad
Minhang Wang$^{3}$ \quad
Jintao Xu$^{3}$ \quad
\\
$^1${School of Information, Renmin University of China, Beijing, China} \quad \\
$^2${Beijing Advanced Innovation Center for Future Blockchain and Privacy Computing,} \\  {Institute of Artificial Intelligence, Beihang University, Beijing, China} \quad \\
$^3${HAOMO.AI, Beijing, China} \quad 
}
\begin{document}
\maketitle
\begin{abstract}
Photo-realistic scene reconstruction from sparse-view, uncalibrated images is highly required in practice. Although some successes have been made, existing methods are either Sparse-View but require accurate camera parameters (i.e., intrinsic and extrinsic), or SfM-free but need densely captured images.
To combine the advantages of both methods while addressing their respective weaknesses, we propose \textbf{Dust to Tower (D2T) }, an accurate and efficient coarse-to-fine framework to optimize 3DGS and image poses simultaneously from sparse and uncalibrated images. Our key idea is to first construct a coarse model efficiently and subsequently refine it using warped and inpainted images at novel viewpoints. 
To do this, we first introduce  a Coarse Construction Module (CCM) which exploits a fast Multi-View Stereo model to initialize a 3D Gaussian Splatting (3DGS) and recover initial camera poses.  To refine the 3D model at novel viewpoints, we propose a Confidence Aware Depth Alignment (CADA) module to refine the coarse depth maps by aligning their confident parts with estimated depths by a Mono-depth model. 
Then, a Warped Image-Guided Inpainting (WIGI) module is proposed to warp the training images to novel viewpoints by the refined depth maps, and inpainting is applied to fulfill the ``holes" in the warped images caused by view-direction changes, providing high-quality supervision to further optimize the 3D model and the camera poses. Extensive experiments and ablation studies demonstrate the validity of D2T and its design choices, achieving state-of-the-art performance in both tasks of novel view synthesis and pose estimation while keeping high efficiency. Codes will be publicly available. 
\end{abstract}    
\section{Introduction}
\label{sec:intro}

\begin{figure}
\centering
	\includegraphics[width=0.9\linewidth]{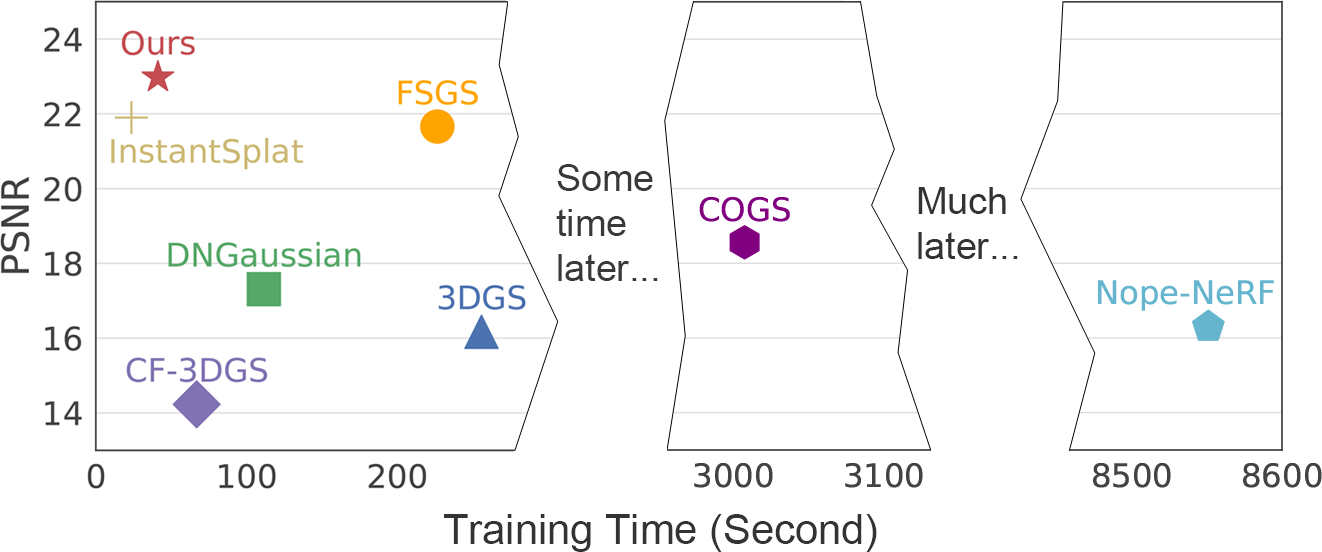}
	\caption{ \textbf{Relationship Between Training Time and PSNR}. We show the training time and PSNR on Tanks and Temples dataset with three input views. Our method is Pareto-optimal on the efficiency-accuracy trade-off when compared to existing baselines.}
	\label{fig:intro}
	\vspace{-3 ex}
\end{figure}
Photo-realistic scene reconstruction is a fundamental task in computer vision and graphics, with applications spanning Augmented Reality (AR)~\cite{SplatLoc}, Virtual Reality (VR)~\cite{VR-GS}, autonomous navigation systems~\cite{AutoSplat}, and beyond~\cite{CaiLWWZYZY24}. Accurate 3D reconstruction enables immersive and interactive experiences, as well as precise spatial understanding in various real-world scenarios. In recent years, significant advancements have been achieved~\cite{NeRF, 3DGS, Mip-NeRF, Mip-Splatting}, pushing the boundaries of what can be achieved in terms of rendering quality and geometric accuracy.

However, despite this progress, existing methods face significant challenges when dealing with real-world conditions. In particular, current approaches can be broadly categorized into two main types: \textbf{Sparse-View} methods and \textbf{SfM-free} methods. Sparse-view methods~\cite{DS-NeRF, DNGaussian, DietNeRF, SinNeRF, FSGS, GeCoNeRF, FewViewGS, ReconFusion} rely on accurately known camera poses and aim to reconstruct scenes from a limited number of images, but they often struggle when pose estimation is inaccurate or unavailable. On the other hand, SfM-free methods~\cite{CF-3DGS, NeRFmm, NoPe-NeRF} jointly estimate camera poses and reconstruct scenes without requiring prior pose information, but most of them depend heavily on densely captured images and perform poorly when only sparse inputs are available. Some works try to reduce the number of views by using long-range information~\cite{COGS, CG-3DGS} or initializing from Multi-View Stereo model~\cite{InstantSplat}, but they suffer from long training time or poor rendering quality at novel viewpoints.
 Both approaches, while effective in specific situations, have limitations in real-world more practical settings where the input images are both sparse and uncalibrated.

To overcome this limitation, we propose a novel \textbf{coarse-to-fine framework} that combines the advantages of both Sparse-View and SfM-free methods, while addressing their respective weaknesses. Our method, named \textbf{Dust to Tower (D2T)}, enables efficient and photo-realistic scene reconstruction from sparse, uncalibrated images (without requiring intrinsic or extrinsic camera parameters) by first constructing a coarse model and then refining it through meticulously warped, and flaw-inpainted images at novel viewpoints. D2T leverages the efficiency of sparse methods while simultaneously solving for camera poses, as done in SfM-free methods, but without the need for densely captured images.

Our framework consists of three key modules: (1) A \textbf{Coarse Construction Module (CCM)} that initializes a 3D Gaussian Splatting (3DGS) model and recovers coarse camera poses using a fast Multi-View Stereo (MVS) model, narrowing the solution space for pose optimization. (2) A \textbf{Confidence Aware Depth Alignment (CADA)} module that refines depth maps by aligning inverse depths from a mono-depth model with coarse depths, ensuring accurate warping of images to novel viewpoints. (3) A \textbf{Warped Image-Guided Inpainting (WIGI)} module that warps input images to novel viewpoints and uses a lightweight inpainting model to fill in the missing regions (called holes), providing high-quality supervision for further refinement of the 3D model and the camera poses.

Extensive experiments demonstrate that \textbf{D2T} achieves state-of-the-art results in both rendering quality and pose estimation accuracy while maintaining high efficiency, as shown in~\cref{fig:intro}. Our contributions can be summarized as follows:

\begin{itemize}
  \item We introduce \textbf{D2T}, a novel coarse-to-fine framework that jointly optimizes 3DGS and camera poses from sparse, uncalibrated images, achieving superior rendering fidelity and efficiency.
  \item We propose \textbf{CADA}, a novel depth alignment technique that enhances warping accuracy by aligning mono-depth predictions with coarse depth estimates.
  \item We develop the \textbf{WIGI} module, which generates multi-view consistent images at novel viewpoints using efficient image warping and inpainting techniques.
  \item Extensive experimental evaluations on multiple datasets demonstrate the state-of-the-art performance of \textbf{D2T} in both novel view synthesis and camera pose estimation, while keeping high efficiency.
\end{itemize}

\section{Related Work}
\label{sec:RelatedWork}

\subsection{Photo-Realistic Scene Reconstruction}

Photo-Realistic Scene Reconstruction is a long-standing problem in computer vision and graphics community~\cite{ChenW93,SeitzD97} and various representations have been explored to address this challenge, including meshes~\cite{Worldsheet,VMesh,ReiserGSVSMBHG24}, light field~\cite{KalantariWR16,WangLWHST18}, planes~\cite{HoiemEH05,Cut-and-Fold} and point clouds~\cite{RPBG,NPBG}.

Recent advancements in neural rendering techniques have highlighted Neural Radiance Field (NeRF)~\cite{NeRF, Nerfstudio} for remarkable ability to produce Photo-Realistic renderings. NeRF uses Multilayer Perceptron (MLP) to represent a 3D scene and images are rendered via volume rendering. However, NeRFs are limited by both training and rendering speed and suffer from artifacts. Substantial follow-ups aim at enhancing either the rendering quality~\cite{Mip-NeRF, Mip-NeRF-360, Zip-NeRF, NeRF++, 0004SC22, NeRF-SR,NeuS} or the time efficiency~\cite{MullerESK22, KiloNeRF, Plenoxels, FastNeRF, LiuGLCT20, 0004SC22}, but achieving both goals simultaneously is still challenging.
More recently, 3D Gaussian Splatting (3DGS)~\cite{3DGS} has shown a significant breakthrough in efficiency and rendering quality for complex real-world scenes. 3DGS represents a scene by a set of anisotropic 3D Gaussian primitives explicitly and enables real-time rendering with a differentiable splatting. 

Despite the success of 3DGS in multiple 3D downstream tasks~\cite{DreamGaussian,LiCLX24,GaussianEditor,SplaTAM,Abdal0SXPKCYW24} and the improvement of rendering quality~\cite{Mip-Splatting, GaussianPro, MS3DGS}, the reliance on dense training images and scene information (\ie precise camera parameters and scene point clouds) severely restricts its practical application. Motivated by the above issues, we design an efficient and effective framework for photo-realistic scene reconstruction from sparse and uncalibrated images.

\subsection{Sparse-View Scene Reconstruction}
Various studies attempt to reconstruct a Photo-Realistic scene using only sparse-view images. Some works~\cite{DS-NeRF, SparseNeRF, DepthRegularizedGS, DNGaussian} use depth priors as additional geometric constraints on input views to mitigate overfitting. But they still underperform due to the lack of supervision from novel viewpoints.
 DietNeRF~\cite{DietNeRF} proposes to supervise at unseen viewpoints by enforcing semantic consistency between training images and rendered images at unseen viewpoints using CLIP~\cite{CLIP}.
  SinNeRF~\cite{SinNeRF} further improves it by using the CLS token from DINO-VIT~\cite{DINO-ViT}. SparseGS~\cite{SparseGS} utilize SDS loss~\cite{DreamFusion} to distill prior knowledge from pre-trained 2D diffusion model~\cite{StableDiffusion}. However, these methods often fail to capture local fine structures. 
  Some methods~\cite{GeCoNeRF, FSGS, FewViewGS} transform input views to unobserved viewpoints to supervise the renderings. Yet, the transformed images have many invisible regions (holes) due to view direction change and  local image expansion. So only limited additional information is introduced, resulting in minor improvement. 
Recent studies use the powerful generative models to produce images at novel viewpoints as pseudo ground truths~\cite{ReconFusion, RECONX, ViewCrafter, LM-Gaussian} or enhance the renderings~\cite{3dgs-enhancer}. But they need complex condition networks to control the generative models, resulting in high computational costs. Other works~\cite{LRM, GS-LRM, Real3D, CRM,M-LRM, MVSplat, pixelSplat, nopo} train networks to directly output 3D model parameters are also related. 

In summary, although achieving some improvement, prior sparse-view works are either hard to provide effective regularization to mitigate the overfitting or suffering from time efficiency. Therefore, we propose the CADA and the WIGI to efficiently generate high-quality supervision at novel viewpoints, resulting in better reconstruction quality.

\subsection{SfM-free Scene Reconstruction} 
Sparse view methods typically rely on precise camera parameters obtained through time-consuming SfM processes (e.g., Colmap~\cite{Colmap}), which often fail with limited images.
To eliminate the dependence on SfM, 
NeRFmm~\cite{NeRFmm} optimize the NeRF and camera parameters jointly, but it is limited to forward-facing scenes. Following NeRFmm, various initiatives have been pursued, including coarse-to-fine positional encoding strategy  for camera poses~\cite{BARF}, progressive optimization strategy~\cite{LocalRF, CF-3DGS}, Gaussian-MLPs for effective pose optimization~\cite{GARF}, and depth-based regularization~\cite{NoPe-NeRF}. While some promising results have been achieved, they generally require dense captured images or/and rough initial poses, and take hours for optimizing a scene. COGS~\cite{COGS} and CG-3DGS~\cite{CG-3DGS} incorporate long-range information using image matching networks~\cite{LoFTR, PRISM} to reduce training views, but they still suffer from training efficiency. 
More recently, InstantSplat~\cite{InstantSplat} uses a pre-trained MVS model~\cite{DUSt3R} to initialize the 3DGS and image poses, and drops the Adaptive Density Control for time efficiency. However, the lack of supervision at novel viewpoints still impedes the reconstruction quality, resulting in blurred artifacts.  

Our work is fundamentally different from existing works in these ways:  1) we propose a coarse-to-fine framework that first constructs a coarse 3D model from sparse and uncalibrated images, then further refine it by warped and inpainted images at novel viewpoints. 2) Comprehensive experiments on both object-level and scene-level datasets show leading performance and efficiency.
\begin{figure*}
\centering
	\includegraphics[width=0.9\textwidth]{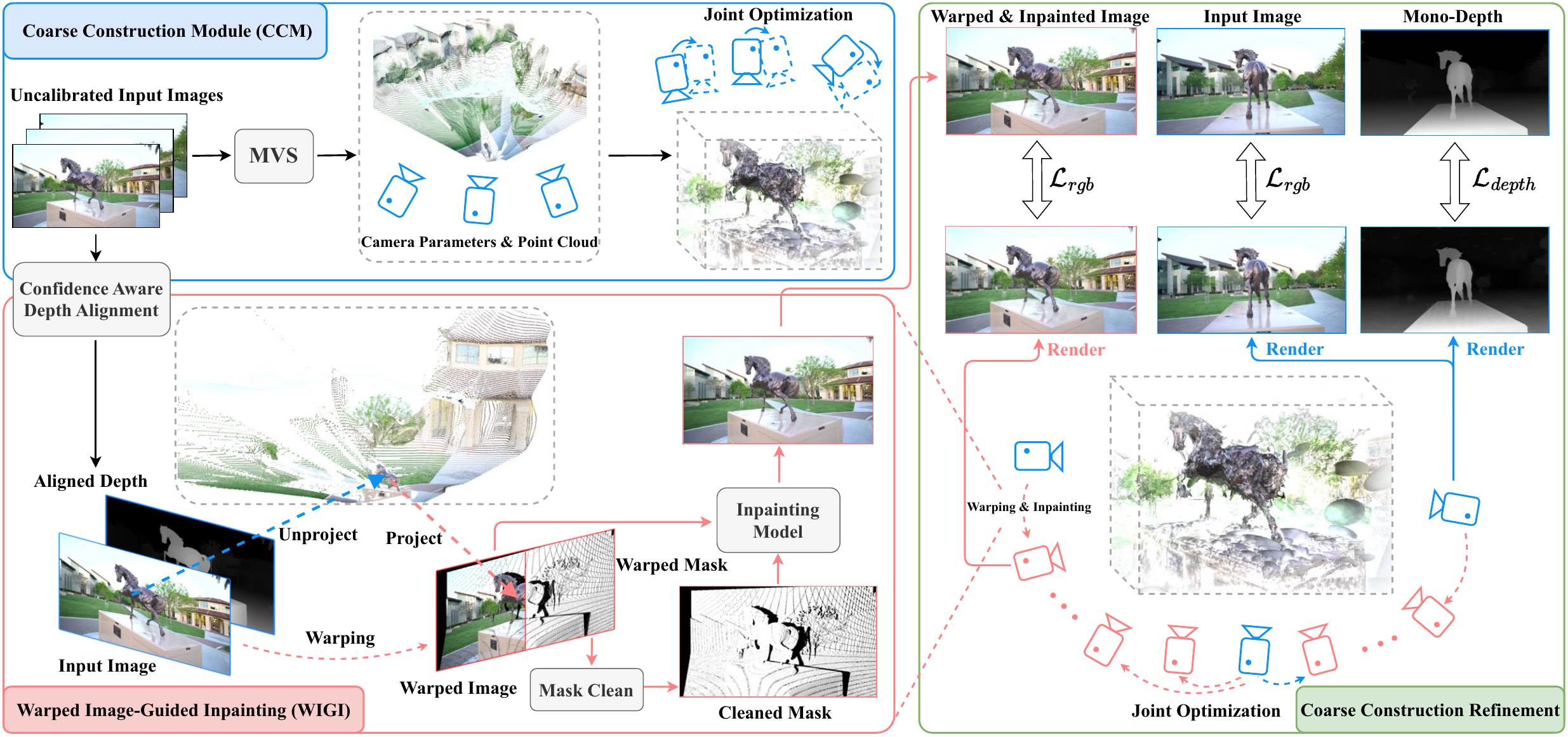}
	\vspace{-1 ex}
	\caption{\textbf{Overview of D2T}. Given sparse-view and uncalibrated images, the Coarse Construction Module (CCM) first employs an efficient MVS method~\cite{DUSt3R} to construct a coarse point cloud and rough camera poses to initialize a 3DGS. The initial 3DGS and poses are optimized simultaneously using the input images for a few steps (\cref{sec:coarseSolution}). To refine the model at novel viewpoints, a Confidence Aware Depth Alignment (CADA) module is proposed to enhance the warping accuracy by aligning relative inverse depth from a SOTA mono-depth model (\cref{sec:CADA}). Then, we propose a Warped Image-Guided Inpainting (WIGI) module to warp input images to unseen viewpoints and inpaint the missing part in the warped images by a lightweight inpainting model (\cref{sec:warping}). Finally, 3DGS and poses are further refined by the inpainted images at novel viewpoints.}
	\label{fig:overview}
	\vspace{-3 ex}
\end{figure*}

\section{Method}
\label{sec:Method}

\subsection{Overview}
Given sparse and uncalibrated images $\mathbf{I} = \{\mathbf{I}_i\}$ for $i \in \{ 1, 2, ..., N \}$, our task is to  reconstruct a photo-realistic 3D model $\mathcal{G}$ for rendering novel views at unseen viewpoints and recover the camera parameters (\ie intrinsic $\mathbf{K}$ and poses $\mathbf{T} = \{\mathbf{T}_i \}$ for $i \in \{ 1, 2, ..., N \}$). We assume a shared $\mathbf{K}$ for all views and use 3DGS as the 3D model representation.  
Our core contribution is to construct a coarse model efficiently and refine it using images at novel viewpoints generated by a warping-and-inpainting paradigm. 
The overview of the proposed method is illustrated in \cref{fig:overview}. We first introduce the Coarse Construction Module in \cref{sec:coarseSolution}. Next, we introduce the Confidence Aware Depth Alignment in \cref{sec:CADA}, which refines the coarse depth maps for accurate warping. \cref{sec:warping} presents the Warped Image-Guided Inpainting module that generates high-quality images at novel viewpoints used to refine the 3D model. Finally, we detail the joint optimization of Poses and 3DGS in \cref{sec:poseOPT}.

\subsection{Coarse Construction Module (CCM)}
\label{sec:coarseSolution}

Instead of the SfM pre-processing, we propose to employ an efficient dense stereo model, \ie DUSt3R~\cite{DUSt3R} to first predict pair-wise pointmaps, then align them into a global point cloud $\chi$ and solve the coarse camera parameters (\ie $\mathbf{K}$ and $\mathbf{T}$). Then, we optimize the poses and 3DGS simultaneously for a few steps to construct a coarse 3DGS.
 Specifically, DUSt3R takes two images $\mathbf{I}_1, \mathbf{I}_2 \in \mathbb{R}^{H\times W \times 3}$ with the longer side not exceeding 512 pixels as input, and output two corresponding pointmaps $X_{1,1}, X_{2,1} \in \mathbb{R}^{H\times W\times 3}$ in the coordinate frame of $\mathbf{I}_1$ and confidence maps $C_{1,1}, C_{2, 1} \in \mathbb{R}^{H \times W}$ in milliseconds. 
 
 We assume a simple pinhole camera model with a centered principal point, then we solve the only unknown parameter in camera intrinsic, \ie the focal length $f^*$ using Weiszfeld algorithm~\cite{Weiszfeld}:
\begin{equation}
\small
f^*=\underset{f}{\arg \min } \sum_{i=0}^W \sum_{j=0}^H C^{i, j}_{1, 1}\left\|\left(i^{\prime}, j^{\prime}\right)-f \frac{\left(X^{i, j, 0}_{1,1}, X^{i, j, 1}_{1,1}\right)}{X^{i, j, 2}_{1,1}}\right\|
\end{equation}
with $i^{\prime}=i-\frac{W}{2} \text { and } j^{\prime}=j-\frac{H}{2}$. As all images share the same camera model, we average the estimated focal lengths of all images to construct the camera intrinsic $\mathbf{K}$. 

As DUSt3R can only deal with image pairs and the scales of the pointmaps are inconsistent between different image pairs, we align multiple pointmaps into a common coordinate frame. Similar to~\cite{DUSt3R}, we first construct a complete connectivity graph $\mathcal{K}_N(\mathcal{V}, \mathcal{E})$ of all the $N$ input images, where vertices $\mathcal{V}$ are images and each edge $e=(n, m) \in \mathcal{E}$ indicates an image pair of $\mathbf{I}_n$ and $\mathbf{I}_m$. We input every image pair $e=(n, m)$ to DUSt3R, resulting the pair-wise predictions $X_{n,n}, X_{m, n}$ and $C_{n,n}, C_{m,n}$. To ensure clarity, we define $X_{n, e}:=X_{n, n}$ and $X_{m, e}:= X_{m,n}$. To align all pointmaps, we introduce a transformation matrix $\mathbf{T}_e$ and scaling factor $\sigma_e$ for each pair $e \in \mathcal{E}$ and optimize the global point cloud $\chi$ as following:
\begin{equation}
\chi^*=\underset{\chi, \mathbf{T}, \sigma}{\arg \min } \sum_{e \in \mathcal{E}} \sum_{v \in \mathcal{V}[e]} \sum_{i=1}^{H W} C^i_{v, e}\left\|\chi^i_v-\sigma_e \mathbf{T}_e X^i_{v, e}\right\|
\end{equation}
Here, $\mathcal{V}[e] = \{n, m\}$ if $e = (n, m)$. The idea is that, for each pair $e$, the same rigid transformation $\mathbf{T}_e$ should align both pointmaps $X_{n,e}$  and $X_{m,e}$ with the world-coordinate pointmaps $\chi_n$ and $\chi_m$. To avoid the trivial solution $ \sigma_e = 0$, for all $e \in \mathcal{E}$, we enforce $\Pi_e \sigma_e = 1$. We directly optimize the global camera poses $\mathbf{T} = \{\mathbf{T}_n \}$ by reparameterization: 
\begin{equation}
\chi_n^{i,j} := \mathbf{T}_n^{-1}h(\mathbf{K}^{-1}[iD_n^{i,j},jD_n^{i,j}, D_n^{i,j}]^{T})
\end{equation}
$D_n$ is the depth map and $h(\cdot)$ denotes the homogeneous coordinates. $i$ and $j$ are the pixel coordinates. The optimization is carried out using standard gradient descent and typically converges within mere seconds on a standard GPU.  
We use the aligned point cloud $\chi$ to initialize the 3DGS and optimize the poses and 3DGS jointly for a few steps to set up a valid skeleton, which will be detailed in \cref{sec:poseOPT}.

\begin{figure}
\centering
	\includegraphics[width=0.9\linewidth]{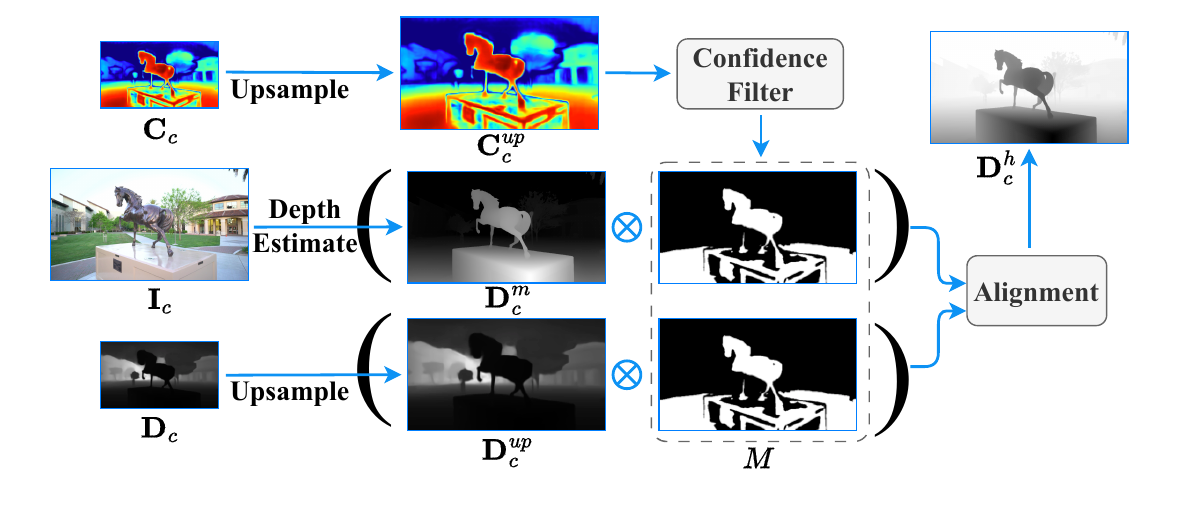}
	\caption{\textbf{Overview of the Confidence Aware Depth Alignment.} We align the mono-depth $\mathbf{D}_c^m$ to the reliable part of the coarse depth $\mathbf{D}_c^{up}$, resulting in high quality depth map $\mathbf{D}_c^h$.}
	\label{fig:depth_align}
\end{figure}

\subsection{Confidence Aware Depth Alignment (CADA)}
\label{sec:CADA}

Warping train images to novel viewpoints relies on precise depth information. Previous methods~\cite{GeCoNeRF, SparseGS, FewViewGS} exploit the rendered depth maps for warping, but the rendered depth maps are inaccurate, since it is the expected depth maps and the 3D model is still rough during training. Directly upsampling the coarse and low-resolution depth maps from CCM may also introduce many noises. To address this issue, we propose using DepthAnything V2 (DAV2)~\cite{DepthAnythingV2} to estimate precise monocular relative inverse depth maps (abbreviated as mono-depth) and align it with the coarse depth maps.

The overview of CADA is illustrated in \cref{fig:depth_align}.
Given a training image $\mathbf{I}_c \in \mathbb{R}^{H_h \times W_h \times 3}$ and corresponding coarse depth map $\mathbf{D}_c \in \mathbb{R}^{H_l \times W_l}$, we predict the mono-depth $\mathbf{D}^{m}_c \in \mathbb{R}^{H_h \times W_h}$ by DAV2 and upsample $\mathbf{D}_c$ to $\mathbf{D}_c^{up} \in \mathbb{R}^{H_h \times W_h}$ via bilinear interpolation. The scale $a$ and shift $b$ between $\mathbf{D}^{m}_c$ and $\mathbf{D}_c^{up}$ are solved by a least squares problem. 
To improve alignment robustness, we use a mask $M$ to retain only the reliable points for alignment, as uncertain depths in $\mathbf{D}_c^{up}$ (\eg in the sky) may cause misalignment.
Specially, we generate a confidence map $C_c \in \mathbb{R}^{H_l \times W_l}$ for $\mathbf{I}_c$ by taking the maximum value at position $(i, j)$ across all confidence maps of $\mathbf{I}_c$ from DUSt3R. Upsample $C_c$ gives $C_c^{up} \in \mathbb{R}^{H_h \times W_h}$, and retaining the top $P$ percent value in $C_c^{up}$ results in  the Mask $M$. Finally, we estimate $a$ and $b$:
\begin{equation}
\scriptsize
(a, b) = \arg\min_{a, b} \sum_{i=1}^{H_h}\sum_{j=1}^{W_h}  M(i,j)\left( \frac{1}{\mathbf{D}_c^{up}(i,j)} - (b + a\mathbf{D}^{m}_c(i,j) \right)^2 
\end{equation}
$\mathbf{D}^m_c$ are aligned to $\mathbf{D}^{up}_c$, resulting in the aligned depth map: $\mathbf{D}^{h}_c = \frac{1}{b + a* \mathbf{D}^{m}_c}$.

\subsection{Warped Image-Guided Inpainting (WIGI)}
\label{sec:warping}

\begin{figure}
\centering
	\includegraphics[width=0.9\linewidth]{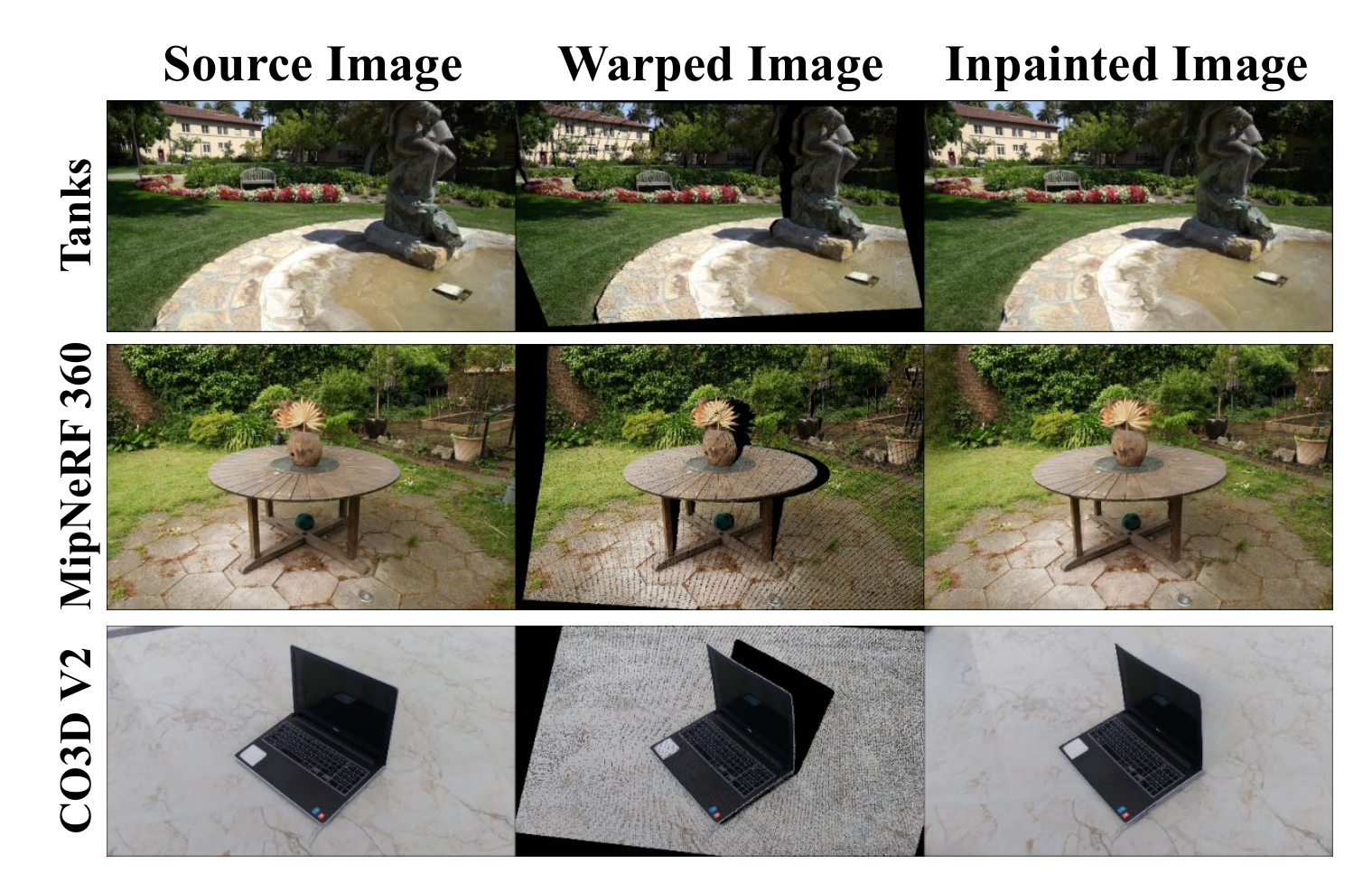}
	\caption{ Visualization of the inpainting results. Tanks means the Tanks and Temples dataset.}
	\label{fig:inpaint_results}
	\vspace{-3 ex}
\end{figure}

To alleviate the overfitting on input views and refine the model at unseen viewpoints,  we propose a novel and efficient method to transform input images to novel viewpoints and fill the missing regions (holes) in warped images by inpainting. Specifically, given $N$ sparse-view images, we sample $K^p * (N-1)$ unseen poses $\mathbf{T}^{p}$ using the B-Spline algorithm for comprehensive coverage of the scenarios. 
For an unseen pose $\mathbf{T}_{k}^{p}$, we warp the closest training image $\mathbf{I}_c$ to $\mathbf{T}_{k}^{p}$ using the aligned depth map $\mathbf{D}^h_c$ of $\mathbf{I}_c$, resulting in the warped image $\mathbf{I}^p_k$ and warped mask $M_k^p$. $\mathbf{I}^p_k$ and $M_k^p$ are used to guide the inpainting model~\cite{LaMa} to eliminate the holes caused by the view direction change, as seen in \cref{fig:inpaint_results}.

\begin{figure}
\centering
	\includegraphics[width=0.9\linewidth]{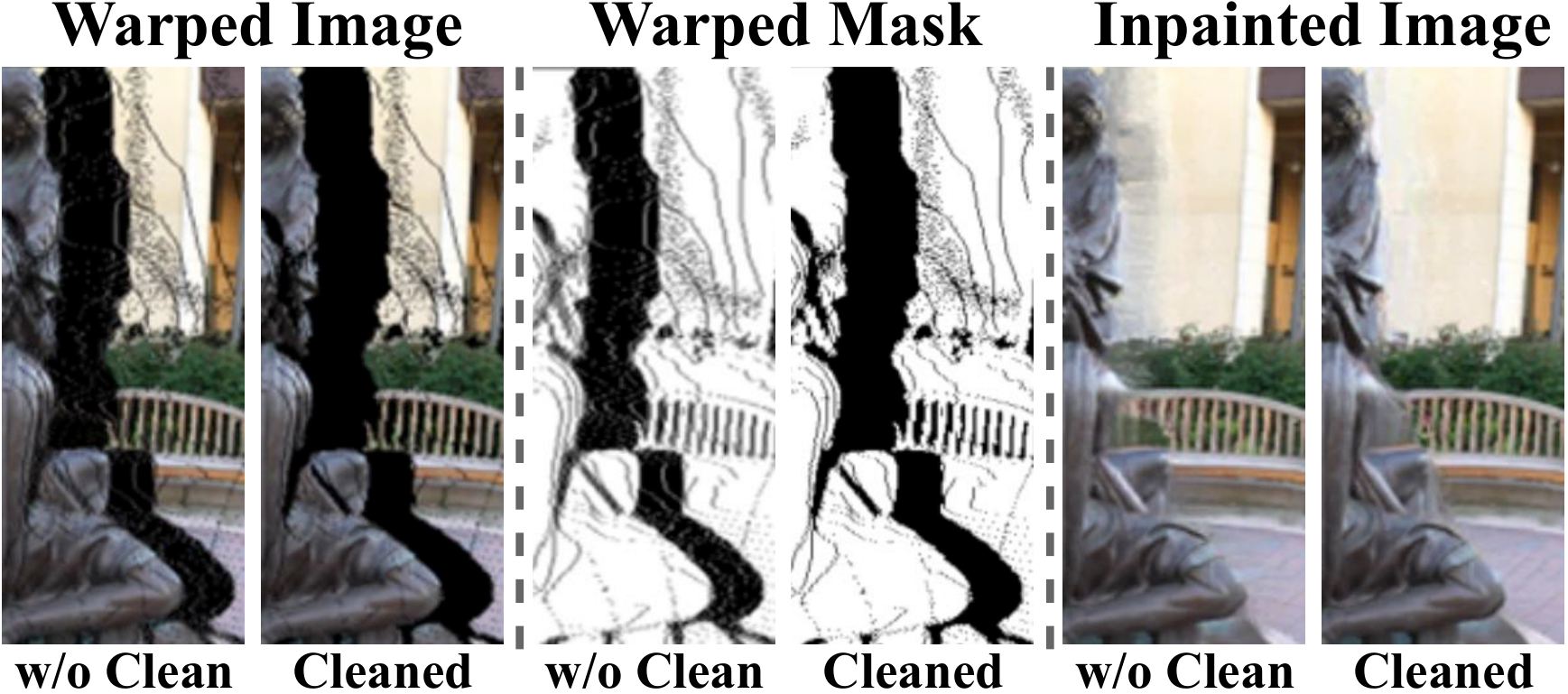}
	\caption{\textbf{Visualization of the Mask Clean}. Without Mask Clean, outliers in the warped image and warped mask can mislead the inpainting model,  resulting in irrational results. The quality of inpainted images are improved after applying Mask Clean.}
	\label{fig:clean_mask}
	\vspace{-3 ex}
\end{figure}

\begin{table*}[]
\setlength{\tabcolsep}{3.5pt}
\centering
\scalebox{0.8}{
\begin{tabular}{cccccccccccccccc}
\hline

\multirow{2}{*}{Category}     & \multirow{2}{*}{Method} & \multicolumn{4}{c}{3 views}                                                &  & \multicolumn{4}{c}{6 views}                                                &  & \multicolumn{4}{c}{12 views}                                               \\ \cline{3-6} \cline{8-11} \cline{13-16} 
                              &                         & PSNR $\uparrow$ & SSIM $\uparrow$ & LPIPS $\downarrow$ & Time $\downarrow$ &  & PSNR $\uparrow$ & SSIM $\uparrow$ & LPIPS $\downarrow$ & Time $\downarrow$ &  & PSNR $\uparrow$ & SSIM $\uparrow$ & LPIPS $\downarrow$ & Time $\downarrow$ \\ \hline
\multirow{3}{*}{Sparse}       & FSGS                    & 21.66           & 0.719           & 0.263              &   3m 47s                &  & {\underline{ 25.70}}     & 0.825           & 0.208              &    3m 41s               &  & {\underline{ 27.78}}     & {\underline{ 0.861}}     & 0.196              &  3m 35s                 \\
                              & DNGaussian              & 17.31           & 0.534           & 0.400              &  1m 52s                 &  & 22.05           & 0.719           & 0.292              &   2m 2s                &  & 24.38           & 0.785           & 0.270              &   2m 7s                \\
                              & 3DGS                    & 16.17           & 0.558           & 0.366              &  4m 16s                 &  & 19.30           & 0.725           & 0.241              &  5m 38s                 &  & 22.48           & 0.808           & 0.179              &  7m 5s                 \\ \hline
\multirow{2}{*}{Unpose}       & CF-3DGS                 & 14.23           & 0.402           & 0.454              &   1m 7s                &  & 15.60           & 0.447           & 0.430              & 2m 10s                   &  & 15.21           & 0.453           & 0.456              &  3m 34s \\
                              & Nope-NeRF               & 16.30           & 0.469           & 0.589              &  2h 22m                 &  & 19.71           & 0.560           & 0.535              &   3h 14m                &  & 21.85           & 0.614           & 0.497              &  5h 17m                 \\ \hline
\multirow{3}{*}{Unconstraint} & InstantSplat            & {\underline{ 21.90}}     & {\underline{ 0.749}}     & {\underline{ 0.218}}        &  \textbf{23s}                 &  & 25.07           & {\underline{ 0.827}}     & {\underline{ 0.150}}        &  \textbf{28s}                 &  & 27.33           & 0.860           & {\underline{ 0.129}}        & \textbf{42s}                 \\
                              & COGS                    & 18.56           & 0.569           & 0.299              &    50m 8s               &  & 22.32           & 0.703           & 0.195              &   1h 19m                &  & 25.60           & 0.810           & 0.128              &    2h 13m               \\
                              & Ours                    & \textbf{23.39}  & \textbf{0.776}  & \textbf{0.164}     &   \underline{41s}               &  & \textbf{26.49}  & \textbf{0.859}  & \textbf{0.124}     &   \underline{49s}                &  & \textbf{27.93}  & \textbf{0.888}  & \textbf{0.113}     &   \underline{1m7s}                \\ \hline
\end{tabular}
}
\vspace{-1 ex}
\caption{ \textbf{NVS results on Tanks and Temples dataset.} The best results are highlighted in bold and the second best results are underlined.}
\label{tab:tanks}
\vspace{-2 ex}
\end{table*}

\paragraph{Image warping with depths}
\label{para:warping}
With the aligned depth map $\mathbf{D}^{h}_c$, we warp the input view $\mathbf{I}_{c}$ at pose $\mathbf{T}_c$ to the unseen viewpoint $\mathbf{T}_k^p$. The transform from $\mathbf{T}_c$  to $\mathbf{T}_k^p$ is given by: $\mathbf{T}_{kc} = (\mathbf{T}_k^p)^{-1} \mathbf{T}_c $. To warp $\mathbf{I}_{c}$ to the unseen viewpoint $\mathbf{T}_k^p$, we first unproject $\mathbf{I}_{c}$ with the aligned depth map $\mathbf{D}^{h}_c$ to 3D space, then reproject it using the transform $\mathbf{T}_{kc}$ and the intrinsic $\mathbf{K}$. Mathematically, the pixel location $p_i$ in $\mathbf{I}_{c}$ transforms to pixel location $p_j$ in the novel view as:
\begin{equation}
p_j = \mathbf{K} \mathbf{T}_{kc}\mathbf{K}^{-1} \mathbf{D}^{h}_c(p_i) p_i
\end{equation}
Then we can map  the pixel colors from $\mathbf{I}_c$ to the warped image $\mathbf{I}^p_k$ by a flow field between $p_i$ and $p_j$ and generate a warp mask $M^p_k$ to indicate the invalid pixels in $\mathbf{I}^p_k$. We use the Z-buffering \cite{Z-buffering} algorithm to resolve the projection ambiguities by retaining the point with the smallest depth when multiple points project to the same pixel location.

\paragraph{Holes Elimination by Inpainting}
After image warping, we have a warped image $\mathbf{I}^p_k$ at unseen viewpoints, which often have holes due to view direction change and local expansion, as shown in  \cref{fig:inpaint_results}.
Previous methods~\cite{GeCoNeRF, SparseGS, SinNeRF,FSGS} mask out the missing region in warped image and only utilize the visible pixels to supervise the model. However, the visible pixels only provided limited additional constraints since they have participated in the optimization of the training views and the missing area is never known and remain unconstrained.  
 To solve this issue, we propose to use the valid parts of the warped image to guide a lightweight inpainting model~\cite{LaMa} $\mathcal{P}$ to generate pixels in the missing regions.  The inpainting model  $\mathcal{P}$ will fill the holes base on the visible content to make the output image natural and coherent. Given the warped image $\mathbf{I}^p_k$ and warped mask $M^p_k$ which indicates areas that need inpainting,  a straightforward way is to directly apply the $\mathbf{I}^p_k$  and $\mathbf{I}^p_k$ for inpainting: $ \hat{\mathbf{I}}^p_k = \mathcal{P}(\mathbf{I}^p_k, M^p_k)$. 
 \vspace{-2 ex}

\paragraph{Mask Clean} However, imperfect and distorted warpings near the edges (\ie discontinuities in depth) can arise from erroneous geometry predicted by DAV2. Such outliers may mislead the model to misunderstand the scene and generate unreasonable content, as shown in~\cref{fig:clean_mask}.
To solve this issue, we propose a simple yet effective method to clean the warped mask $M^p_k$: $\hat{M}^p_k = \mathcal{F} (M^p_k)$.  Specifically, a point is deemed an outlier if it has less than $\frac{w \times w}{2}$ neighboring points within its $w \times w$ neighborhood. The  final inpainted images are obtained as $\hat{\mathbf{I}}^p_k = \mathcal{P}(\mathbf{I}^p_k, \hat{M}^p_k)$, where $\hat{M}^p_k$ indicates the valid and visible points. We can then use $\hat{\mathbf{I}}^p_k$ at the novel viewpoints $\mathbf{T}_k^p$ to refine the 3DGS model.

Comparing to concurrent works~\cite{ReconFusion, RECONX, ViewCrafter, LM-Gaussian} which  directly generate complete images at novel viewpoints using the pre-trained generative models~\cite{DynamiCrafter, StableDiffusion, ModelScope}, WIGI offers more robust multi-view consistency through explicit geometric correspondences and greater efficiency.

\begin{table*}[]
\setlength{\tabcolsep}{3.5pt}
\centering
\scalebox{0.8}{
\begin{tabular}{cccccccccccccccc}
\hline
\multirow{2}{*}{Category}     & \multirow{2}{*}{Method} & \multicolumn{4}{c}{3 views}                                                &  & \multicolumn{4}{c}{6 views}                                                &  & \multicolumn{4}{c}{12 views}                                               \\ \cline{3-6} \cline{8-11} \cline{13-16} 
                              &                         & PSNR $\uparrow$ & SSIM $\uparrow$ & LPIPS $\downarrow$ & Time $\downarrow$ &  & PSNR $\uparrow$ & SSIM $\uparrow$ & LPIPS $\downarrow$ & Time $\downarrow$ &  & PSNR $\uparrow$ & SSIM $\uparrow$ & LPIPS $\downarrow$ & Time $\downarrow$ \\ \hline
\multirow{3}{*}{Sparse}       & FSGS                    & 12.33           & 0.261           & 0.637              &  3m 17s                 &  & 14.18           & 0.339           & 0.584              &   3m 18s                &  & \underline{17.27}           & 0.477           & 0.507              &  3m 28s                 \\
                              & DNGaussian              & 11.37           & 0.235           & 0.694              &  2m 45s                 &  & 13.19           & 0.348           & 0.639              &  2m 52s                 &  & 14.43           & 0.402           & 0.622              &   2m 57s                \\
                              & 3DGS                    & 11.56           & 0.188           & 0.625              &  6m 41s                 &  & 13.06           & 0.261           & 0.575              &   7m 6s                &  & 14.88           & \underline{0.493}           & \underline{0.375}              &  7m 38s                 \\ \hline
\multirow{2}{*}{Unpose}       & CF-3DGS                 & 12.70           & 0.227           & 0.594              &   1m 10s                &  & 13.37           & 0.230           & 0.590              &   2m 19s                &  & 13.96           & 0.260           & 0.602              &  4m 2s                 \\
                              & Nope-NeRF               & \underline{14.43}           & \underline{0.304}           & 0.702              & 2h 12m                  &  & \underline{15.86}           & 0.351           & 0.685              & 3h 7m                  &  & 17.02           & 0.384           & 0.662              &  5h 1m                 \\ \hline
\multirow{3}{*}{Unconstraint} & InstantSplat            & 13.77           & 0.285           & \underline{0.551}              &   \textbf{23s}                &  & 15.34           & \underline{0.399}           & \underline{0.455}              &    \textbf{30s}               &  & 17.09           & 0.456           & 0.493              &   \textbf{49s}                \\
                              & COGS                    & 12.48           & 0.204           & 0.593              & 1h 7m                  &  & 13.60           & 0.257           & 0.557              &  1h 44m                 &  & 15.72           & 0.342           & 0.480              & 2h 28m                  \\
                              & Ours                    & \textbf{14.99}           & \textbf{0.331}           & \textbf{0.524}              &  \underline{45s}                 &  & \textbf{17.80}           & \textbf{0.443}           & \textbf{0.396}              &   \underline{53s}                &  & \textbf{19.73}           & \textbf{0.501}           & \textbf{0.354}              &   \underline{1m19s}                \\ \hline
\end{tabular}
}
\vspace{-1 ex}
\caption{ \textbf{NVS results on MipNeRF360 dataset.} The best results are highlighted in bold and the second best results are underlined.}
\label{tab:mipnerf}
\vspace{-3 ex}
\end{table*}

\begin{table*}[]
\setlength{\tabcolsep}{3.5pt}
\centering
\scalebox{0.8}{
\begin{tabular}{cccccccccccccccc}
\hline
\multirow{2}{*}{Category}     & \multirow{2}{*}{Method} & \multicolumn{4}{c}{3 views}                                                &  & \multicolumn{4}{c}{6 views}                                                &  & \multicolumn{4}{c}{12 views}                                               \\ \cline{3-6} \cline{8-11} \cline{13-16} 
                              &                         & PSNR $\uparrow$ & SSIM $\uparrow$ & LPIPS $\downarrow$ & Time $\downarrow$ &  & PSNR $\uparrow$ & SSIM $\uparrow$ & LPIPS $\downarrow$ & Time $\downarrow$ &  & PSNR $\uparrow$ & SSIM $\uparrow$ & LPIPS $\downarrow$ & Time $\downarrow$ \\ \hline
\multirow{3}{*}{Sparse}       & FSGS                    & 17.99           & 0.731           & 0.438              &   4m 47s                &  & 20.35           & 0.770           & 0.395              &   5m 8s                &  & 22.32           & 0.802           & 0.367              &  5m 30s                 \\
                              & DNGaussian              & 15.17           & 0.703           & 0.476              &   5m 3s                &  & 17.04           & 0.738           & 0.441              &   5m 26s               &  & 20.31           & 0.796           & 0.389              &   5m 30s                \\
                              & 3DGS                    & 16.10           & 0.662           & 0.458              & 8m 29s                  &  & 18.59           & 0.722           & 0.419              &   9m 52s                &  & 19.96           & 0.762           & 0.387              &   11m 11s                \\ \hline
Unpose                        & CF-3DGS                 & 16.27           & 0.713           & 0.445              &   3m 30s                &  & 16.87           & 0.717           & 0.449              &  5m 53s                 &  & 17.06           & 0.731           & 0.457              &   8m 39s                \\ \hline
\multirow{3}{*}{Unconstraint} & InstantSplat            & \underline{18.15}           & \underline{0.741}           & \underline{0.362}              &  \textbf{30s}                 &  & \underline{22.24}           & \underline{0.826}           & \underline{0.283}              & \textbf{37s}                   &  & \underline{25.75}           & \underline{0.869}           & \underline{0.242}              &  \textbf{55s}                 \\
                              & COGS                    & 16.10           & 0.669           & 0.455              & 14m 39s                  &  & 17.17           & 0.696           & 0.435              &  26m 4s                 &  & 18.07           & 0.726           & 0.410              & 42m 3s                  \\
                              & Ours                    & \textbf{19.79}           & \textbf{0.771}           & \textbf{0.345}              &  \underline{59s}                 &  & \textbf{24.74}           & \textbf{0.854}           & \textbf{0.250}              &   \underline{1m 12s}                &  & \textbf{27.08}           & \textbf{0.876}           & \textbf{0.225}              &   \underline{1m 36s}                \\ \hline
\end{tabular}
}
\vspace{-1 ex}
\caption{ \textbf{NVS results on CO3D V2 dataset.} The best results are highlighted in bold and the second best results are underlined.}
\label{tab:co3d}
\vspace{-2 ex}
\end{table*}

\begin{figure*}
\centering
	\includegraphics[width=0.9\linewidth]{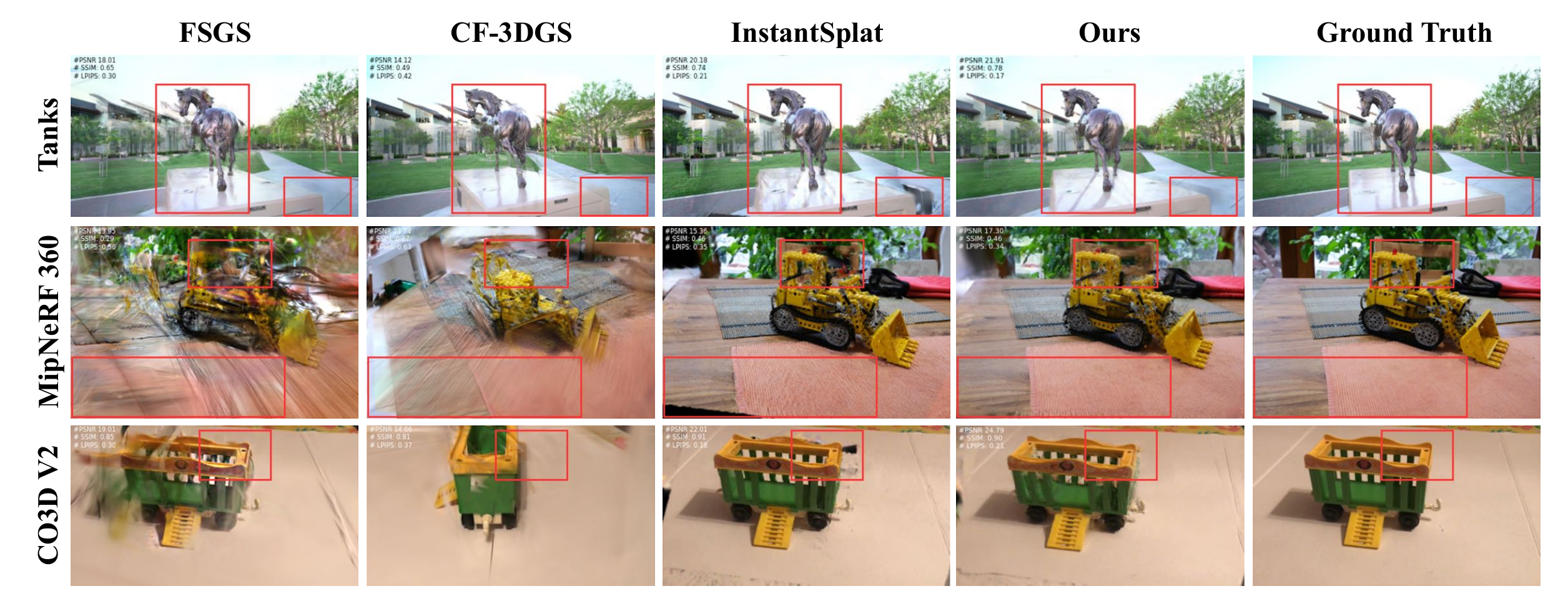}
	\caption{\textbf{Qualitative Results of Novel View Synthesis with three training views.} Our method produces more realistic rendering results with fewer artifacts than others, especially in the occluded regions. The evaluation metrics against ground truth are plot at the top and left of the images. Please zoom in for detail.}
	\label{fig:nvs_results}
	\vspace{-3 ex}
\end{figure*}

\subsection{Joint Optimization of Poses and 3DGS}
\label{sec:poseOPT}
In this section, we first discuss the optimization of camera poses, then we present the loss function used for training. Finally, we detail the coarse-to-fine optimization strategy.

\paragraph{Pose Optimization} We optimize the camera poses by adding tiny perturbations $\Delta \mathbf{T}$ to the initial poses $\mathbf{T}$. The images and depths rendered at perturbed poses are expressed by: $\tilde{ \mathbf{I}},\tilde{ \mathbf{D}} =  \mathcal{R}(\mathcal{G}, \Delta \mathbf{T} \times \mathbf{T})$, where $\mathcal{R}(\cdot)$ refers to the rendering process.
We update $\Delta \mathbf{T}$ using stochastic gradient descent alongside scene optimization to refine the poses of training images.

\paragraph{Loss Function} 
Following 3DGS~\cite{3DGS}, we train the 3DGS by minimizing the difference between training images $\mathbf{I}$ and rendering images $\tilde{\mathbf{I}}$ using stochastic gradient descent:
\begin{equation}
\mathcal{L}_{\text {rgb}}=(1-\lambda) \mathcal{L}_1(\tilde{\mathbf{I}}, \mathbf{I})+\lambda \mathcal{L}_{D-S S I M}(\tilde{\mathbf{I}}, \mathbf{I})
\label{eq:rgb}
\end{equation}

To achieve coherent geometry, we introduce a geometry prior to the optimization. We use the predicted high-quality mono-depths $\mathbf{D}^m$ as a depth prior. 
To avoid scale discrepancies, we convert the rendered depth maps $\tilde{ \mathbf{D}}$ into inverse depth maps and use a loose constraint, 
Pearson correlation which measures the consistency between two variables for supervision. The loss $\mathcal{L}_{depth}$ is defined as follows:
\begin{equation}
	\mathcal{L}_{depth} =1- \rho (\frac{1}{\tilde{ \mathbf{D}}}, \mathbf{D}^m),
\rho(X, Y) = \frac{cov(X, Y)}{\sigma_{X}\sigma_{Y}}
\end{equation}
$cov(\cdot)$ denotes covariance. $\sigma_{X}, \sigma_{Y}$ are standard deviation.

\vspace{-2 ex}
\paragraph{Coarse-to-Fine Optimization Strategy}
As noted in \cref{para:warping}, the warping process requires accurate camera poses. Directly using the coarse poses from DUSt3R may lead to declined performance. Thus, we propose a Coarse-to-Fine Optimization Strategy for more robust training.

In the coarse stage, we first train the coarse 3DGS model and poses from DUSt3R using only the training views for $k_1$ steps. The loss function in the coarse stage is: 
\begin{equation}
\mathcal{L}_c = \mathcal{L}_{rgb}(\mathbf{I}, \tilde{\mathbf{I}}) + \lambda_d \mathcal{L}_{depth}(\tilde{\mathbf{D}}, \mathbf{D}^m).
\end{equation}

In the fine stage, we use the updated input image poses to perform WIGI (\ie \cref{sec:warping}), resulting in $K^p \times (N-1)$ images at novel viewpoints. We supervise the corresponding rendered images by $\mathcal{L}_{rgb}$ as an additional regularization. We refine the 3DGS $K_2$ steps with the fine stage loss:

\begin{equation}
\mathcal{L}_f = \mathcal{L}_c + \lambda_{pseudo} \mathcal{L}_{rgb}(\hat{\mathbf{I}}^p, \tilde{ \mathbf{I}}^p)
\end{equation}
where $ \tilde{\mathbf{I}}^p$ means the rendered images and $\hat{\mathbf{I}}^p$ is the warped and inpainted images  at unseen viewpoints.
\section{Experiments}
\label{sec:Exp}

\subsection{Experiment Setup}

\paragraph{Datasets} We conduct extensive experiments on three datasets, encompassing both object and scene reconstruction scenarios. \textbf{(1) Tanks and Temples}~\cite{Tanks} dataset is widely used. We use the same eight scenes for evaluation as used in~\cite{NoPe-NeRF, COGS}. For each scene, we uniformly sample 24 images and then uniformly select 12 images from it (including the first and last images) as the training set, $N$ sparse training views are sampled from the training set.
 The remaining 12 images are used for test. \textbf{(2) MipNeRF360} dataset~\cite{Mip-NeRF-360} features  complex central objects or areas against a detailed background. We use the published seven scenes for evaluation. We uniformly sample 24 images  from the first 48 frames and split the train/test set similarly. \textbf{(3) CO3D V2} dataset~\cite{CO3D} consists of thousands of videos with 360-degree coverage around central objects. We randomly select 8 videos of different objects and follow the same protocol to split the train/test set as the Tanks and Temples dataset.

\begin{table*}[]
\setlength{\tabcolsep}{3.5pt}
\centering
\scalebox{0.78}{
\begin{tabular}{ccccccccccccccccccccc}
\hline
\multirow{2}{*}{views}    & \multirow{2}{*}{Dataset} & \multicolumn{3}{c}{Ours}                &  & \multicolumn{3}{c}{COGS}                &  & \multicolumn{3}{c}{CF-3DGS}             &  & \multicolumn{3}{c}{Nope-NeRF}           &  & \multicolumn{3}{c}{InstantSplat}        \\ \cline{3-5} \cline{7-9} \cline{11-13} \cline{15-17} \cline{19-21} 
                          &                          & $\text{RPE}_r$ & $\text{RPE}_t$ & ATE   &  & $\text{RPE}_r$ & $\text{RPE}_t$ & ATE   &  & $\text{RPE}_r$ & $\text{RPE}_t$ & ATE   &  & $\text{RPE}_r$ & $\text{RPE}_t$ & ATE   &  & $\text{RPE}_r$ & $\text{RPE}_t$ & ATE   \\ \hline
\multirow{3}{*}{3 views}  & Tanks                    & \textbf{0.416}          & \textbf{0.807}          & \textbf{0.003} &  & \underline{1.105}          & \underline{3.809}          & \underline{0.017} &  & 21.369         & 73.933         & 0.237 &  & 19.785         & 50.972         & 0.155 &  & 1.800          & 12.221         & 0.025 \\
                          & Mipnerf                  & \textbf{2.642}          & \textbf{3.269}          & \textbf{0.009} &  & 73.903         & \underline{86.434}         & \underline{0.187} &  & 82.720         & 110.053        & 0.348 &  & 81.647         & 128.994        & 0.323 &  & \underline{23.104}         & 96.475         & 0.267 \\
                          & CO3D                     & \textbf{13.370}         & \textbf{11.731}         & \textbf{0.026} &  & 121.102        & \underline{100.196}        & 0.369 &  & 121.475        & 105.898        & \underline{0.280} &  & -              & -              & -     &  & \underline{103.141}        & 118.120        & 0.323 \\ \hline
\multirow{3}{*}{6 views}  & Tanks                    & \textbf{0.222}          & \textbf{0.364}          & \textbf{0.003} &  & \underline{0.258}          & \underline{0.588}          & \underline{0.004} &  & 7.412          & 23.579         & 0.172 &  & 6.002          & 15.278         & 0.114 &  & 1.221          & 5.459          & 0.025 \\
                          & Mipnerf                  & \textbf{7.146}          & \textbf{5.940}          & \textbf{0.022} &  & 53.527         & \underline{38.996}         & \underline{0.205} &  & 61.405         & 49.423         & 0.303 &  & 61.791         & 107.246        & 0.547 &  & \underline{20.655}         & 55.885         & 0.281 \\
                          & CO3D                     & \textbf{5.569}          & \textbf{7.203}          & \textbf{0.054} &  & 64.521         & 42.951         & 0.320 &  & \underline{50.596}         & \underline{36.459}         & \underline{0.248} &  & -              & -              & -     &  & 97.642         & 50.471         & 0.280 \\ \hline
\multirow{3}{*}{12 views} & Tanks                    & \underline{0.490}          & \textbf{0.194}          & \textbf{0.002} &  & \textbf{0.119}          & \underline{0.299}          & \underline{0.004} &  & 2.553          & 7.075          & 0.099 &  & 2.162          & 7.147          & 0.111 &  & 0.886          & 2.128          & 0.015 \\
                          & Mipnerf                  & \textbf{1.896}          & \textbf{1.613}          & \textbf{0.012} &  & 23.839         & \underline{14.940}         & \underline{0.149} &  & 29.861         & 19.974         & 0.216 &  & 29.865         & 58.931         & 0.541 &  & \underline{13.632}         & 26.620         & 0.207 \\
                          & CO3D                     & \textbf{2.787}          & \textbf{2.810}          & \textbf{0.039} &  & 36.361         & 18.785         & 0.203 &  & \underline{22.164}         & \underline{13.539}         & \underline{0.171} &  & -              & -              & -     &  & 22.644         & 45.430         & 0.207 \\ \hline
\end{tabular}
}
\vspace{-1 ex}
\caption{\textbf{Pose Estimation Evaluation results.} The $\text{RPE}_t$ is scaled by 100, the $\text{RPE}_r$ is in degree and the $\text{ATE}$ is in the ground truth scale. The best results are highlighted in bold and the second best results are underlined.}
\vspace{-2 ex}
\label{tab:pose}
\end{table*}

\vspace{-3 ex}
\paragraph{Metrics}
We evaluate the tasks of \textbf{Novel View Syntheses} (NVS) and \textbf{Camera Pose Estimation}. For NVS, we report Peak Signal-to-Noise Ratio (PSNR), Structural Similarity Index Measure (SSIM)~\cite{SSIM}, and Learned Perceptual Image Patch Similarity (LPIPS)~\cite{LPIPS}. For camera pose estimation, we align the estimated poses with the ground truth using the Umeyama algorithm~\cite{Umeyama}  and report the Absolute Trajectory Error (ATE) and Relative Pose Error (RPE) following~\cite{NoPe-NeRF, CF-3DGS}. For Tanks and Temples and MipNeRF360 datasets, we consider Colmap~\cite{Colmap} poses as the ground truth. CO3D V2 provides its own ground truth poses.
\vspace{-2 ex}

\paragraph{Baselines}
We compare our method with three types of methods: 1) \textbf{Sparse methods}~\cite{FSGS, DNGaussian}, which reconstruct from sparse views but requires pre-computing camera parameters. We use ground truth parameters to train them. Vanilla 3DGS~\cite{3DGS} is also evaluated as a baseline. 2) \textbf{Unpose methods}~\cite{CF-3DGS, NoPe-NeRF} that reconstruct from uncalibrated images but requires dense captured images; 3) \textbf{Unconstraint methods}~\cite{InstantSplat, COGS} that reconstruct from sparse and uncalibrated images. We use their official codes and default configurations to train all methods and report the performance. 

\vspace{-2 ex}
\paragraph{Implementation Details}
We develop our entire framework based on PyTorch~\cite{PyTorch}  and use gsplat~\cite{gsplat} for differentiable rendering. We follow the default optimization parameters of 3DGS~\cite{3DGS} unless otherwise specified. All experiments are conducted on an Nvidia 4090 GPU. We construct the coarse solution with 300 iterations and refine the 3D model for another 1700 iterations. We set $K^p=18, 6, 4$ for $3, 6, 12$ training view settings, respectively. And we set $\lambda=0.1, \lambda_{d} = 0.5, \lambda_{pseudo} = 0.3$ for all experiments. More details are provided in the supplementary material.

\vspace{-2 ex}
\paragraph{Test View Pose Alignment}
To evaluate the Novel View Synthesis performance, we need to render images at the same viewpoints as the test images. But the poses of the  test images are unknown, we need to localize the test images in the trained 3DGS model. Inspired by NeRFmm~\cite{NeRFmm}, we first freeze the trained 3DGS model, and then optimize the camera poses of the test views via stochastic gradient descent. The optimization target is to minimize the photometric loss between the rendered images and the test views. We optimize 200 steps for each test view. 

\subsection{Evaluation on Novel View Synthesis}

\cref{tab:tanks}, \cref{tab:mipnerf}, \cref{tab:co3d}  report the averaged results and training time on Tanks and Temples, MipNeRF360 and CO3D V2 datasets, respectively. Our method consistently outperforms all baselines across all the datasets in the PSNR, SSIM, and LPIPS scores and achieves  top-tier training efficiency, which strongly proves the effectiveness and robustness of our method. The qualitative results are shown in Fig~\ref{fig:nvs_results}. InstantSplat fails to reconstruct the occluded regions at novel viewpoints, resulting in holes. CF-3DGS and FSGS suffer from blurred artifacts due to the overfitting issue. Our method successfully reconstructs the missing part at novel viewpoints, and the rendered images are more clearer and with fewer artifacts.

\subsection{Evaluation on Camera Pose Estimation}
We report the pose estimation results of three datasets in \cref{tab:pose}. Our approach achieves new state-of-the-art performance for most evaluation metrics, demonstrating the effectiveness of our method. Notably, our method  is more accurate than the compared methods by an order of magnitude in ATE, and the $\text{RPE}_r$ and $\text{RPE}_t$ are also significantly improved across all scenarios. CF-3DGS and NoPE-NeRF rely on dense input and struggle to estimate accurate camera poses in sparse-view settings.

\subsection{Ablation Study}
To fully understand the design choices in our method,  we conduct  ablation studies on the Tanks and Temples dataset with three training views. 
The results are shown in \cref{tab:abl}.

\vspace{-1 ex}
\paragraph{Impact of CADA} The variant \textbf{(a)} removes the CADA module and directly uses the upsampled coarse depth maps for warping. It makes the warping process inaccurate, resulting in declined reconstruction quality.

\vspace{-1 ex}
\paragraph{Effectiveness of WIGI} As shown in \cref{tab:abl} \textbf{(b)}, omitting WIGI results in a significant performance drop, since the unseen regions lack supervision. Besides,  the variant \textbf{(c)} ablates the inpainting step and directly uses the warped images for supervision, leading to noticeably declined results. It validates the necessity of the inpainting process, which distinguishes our method from prior warping-based methods. The Mask Clean also contributes to the rendering quality,  as evidenced by the results of variant \textbf{(d)}.

\vspace{-1 ex}
\paragraph{Joint Optimization of Poses and 3DGS Matters} Variant \textbf{(e)} disables the optimization of camera poses, and uses the initial camera poses from DUSt3R to perform WIGI and train the 3DGS. As shown in \cref{tab:abl}, it leads to declined pose accuracy, the rendering quality is also impeded due to the inaccurate camera poses. The Coarse-to-Fine Optimization Strategy is also critical to the reconstruction quality, proved by the results of variant \textbf{(f)}.

\begin{table}[]
\centering
\scalebox{0.8}{
\begin{tabular}{lcccc}
\hline
Variant & PSNR $\uparrow$ & SSIM $\uparrow$ & $\text{RPE}_t$ $\downarrow$  & $\text{RPE}_r$ $\downarrow$ \\ \hline
\textbf{(a)} w/o CADA    & 22.89 & 0.7632 & 0.8253      & 0.4189     \\
\textbf{(b)} w/o WIGI    & 22.18 & 0.7511 & 0.8101      & 0.4175     \\
\textbf{(c)} w/o Inpainting    & 22.29 & 0.7550 & 0.8223      & 0.4162      \\
\textbf{(d)} w/o Mask Clean    & 22.90 & 0.7666 & 0.8122      & 0.4193     \\ 
\textbf{(e)} w/o  Joint Optimization  & 22.69 & 0.7547 & 1.4681      & 0.4526     \\ 
\textbf{(f)} w/o Coarse-to-Fine Strategy    & 22.85 & 0.7621 & 1.0384      & 0.4261  \\ \hline
\textbf{Full}  & \textbf{23.39} & \textbf{0.7762} & \textbf{0.8068}      & \textbf{0.4159}     \\ \hline
\end{tabular}
}
\vspace{-1 ex}
\caption{Ablation study on Tanks and Temples dataset with three training views.}
\label{tab:abl}
\vspace{-3 ex}
\end{table}

\begin{figure}[ht]
\centering
\begin{minipage}[c]{0.48\linewidth}
    \centering
    \includegraphics[width=\linewidth]{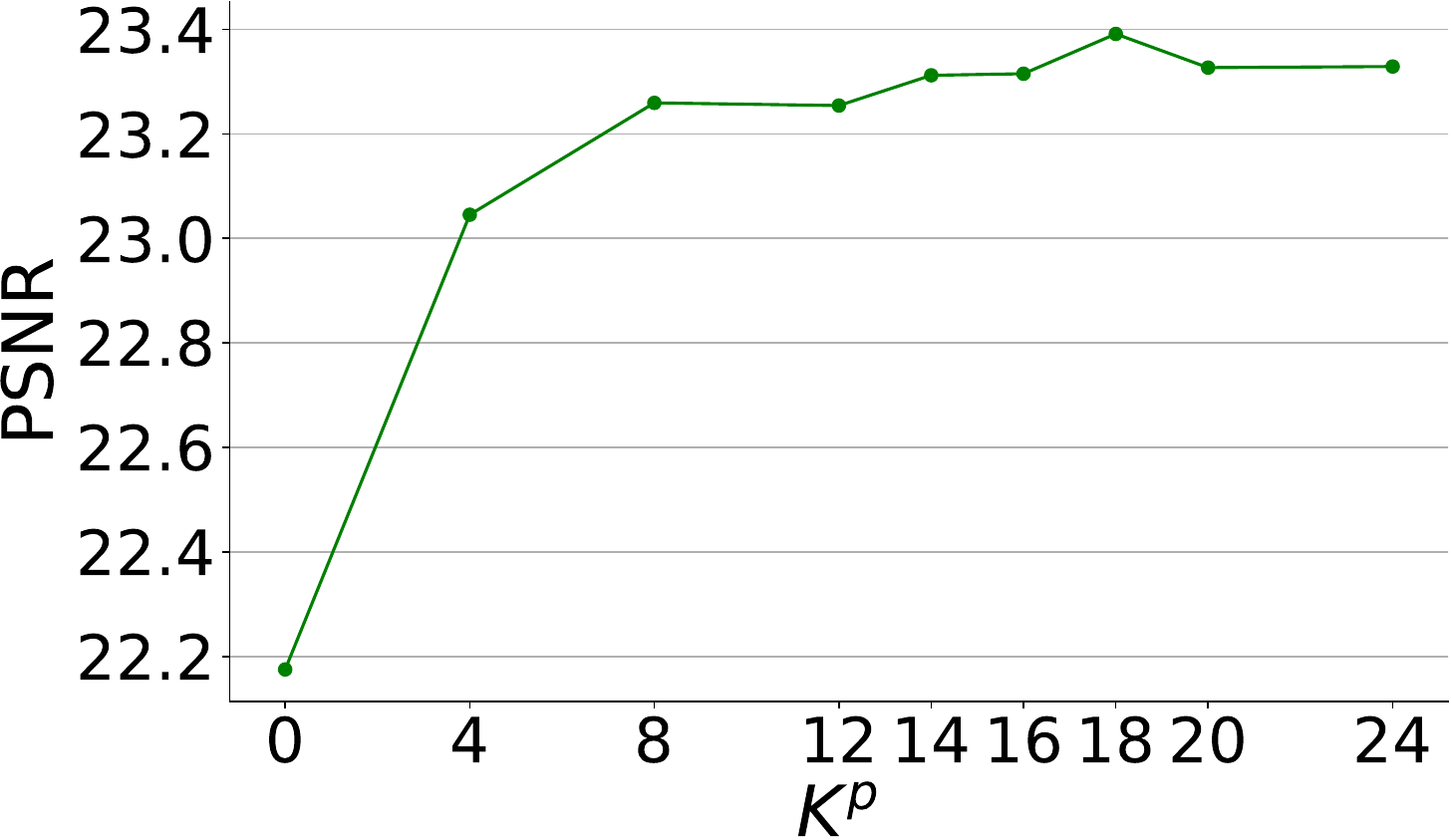}
    \caption{Impact of the number of inpainting views}
    \label{fig:num-pviews}
\end{minipage}%
\hfill
\begin{minipage}[c]{0.48\linewidth}
    \centering
    \includegraphics[width=\linewidth]{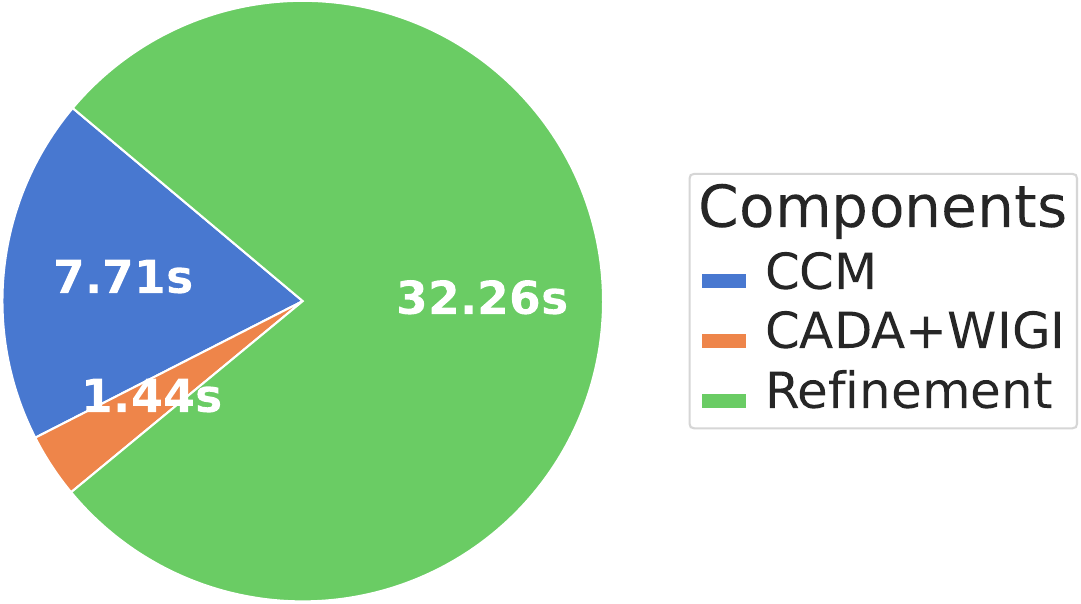}
    \caption{Time Analyses of Different Components.}
    \label{fig:time}
\end{minipage}
\vspace{-3 ex}
\end{figure}

\paragraph{Impact of the number of inpainting views}
We present PSNR results for varying numbers of inpainting views on the Tanks and Temples dataset with three input views, as demonstrated in \cref{fig:num-pviews}. Initially, the PSNR significantly improves due to the inclusion of warped and inpainted images. As $K^p$  increases, the improvement in PSNR becomes marginal.

\vspace{-1 ex}
\paragraph{Time Analyses}
\cref{fig:time} shows the average time consumption of different modules, which is tested on the Tanks and Temples dataset with three training views. We set $K^p=18$, resulting in 36 warped and inpainted images in total. Our proposed CCM can construct a coarse 3D model and recover initial camera poses in mere seconds.  It only takes $1.44s$ to generate 36 warped and inpainted images, proving the efficiency of the proposed CADA and WIGI.

\section{Conclusion}
\label{sec:Conclusion}
In this paper, we propose D2T, a novel coarse-to-fine framework that aims to use sparse and uncalibrated images for photo-realistic scene reconstruction. Starting from sparse and uncalibrated images, D2T first constructs a coarse solution efficiently by CCM. To refine the 3D model at novel viewpoints, we propose CADA and WIGI to generate images at novel viewpoints by warping and inpainting, which is proven effective and efficient in enhancing the rendering quality at novel viewpoints. Extensive experimental results on three datasets show that D2T achieves state-of-the-art results in both NVS and pose estimation with high efficiency.
However, the alignment of a global point cloud limited D2T to reconstruct large-scale scenes with hundreds of input images. An incremental alignment paradigm will be explored in future research.
{
    \small
    \bibliographystyle{ieeenat_fullname}
    \bibliography{main}
}


\end{document}